\PassOptionsToPackage{xcolor}{pdftex,dvipsnames}
\documentclass[11pt,a4paper]{article}
\usepackage[hyperref]{naaclhlt2018}
\usepackage{times}
\usepackage{latexsym}

\usepackage{url}

\usepackage{amsmath}
\usepackage{amssymb}
\usepackage{booktabs}
\usepackage{xspace}
\usepackage{enumitem}
\usepackage{xargs}
\usepackage{xr}
\usepackage{contour}
\usepackage{titlesec}
\usepackage{multirow}
\usepackage{arydshln}
\usepackage{bm}
\usepackage[hang,flushmargin]{footmisc}
\usepackage[colorinlistoftodos,prependcaption,textsize=scriptsize]{todonotes}
\newcommandx{\unsure}[2][1=]{\todo[linecolor=red,backgroundcolor=red!25,bordercolor=red,#1]{#2}}
\newcommandx{\change}[2][1=]{\todo[linecolor=blue,backgroundcolor=blue!25,bordercolor=blue,#1]{#2}}
\newcommandx{\info}[2][1=]{\todo[linecolor=OliveGreen,backgroundcolor=OliveGreen!25,bordercolor=OliveGreen,#1]{#2}}
\newcommandx{\improvement}[2][1=]{\todo[linecolor=Plum,backgroundcolor=Plum!25,bordercolor=Plum,#1]{#2}}
\newcommandx{\thiswillnotshow}[2][1=]{\todo[disable,#1]{#2}}
\makeatletter
\g@addto@macro\normalsize{%
  \setlength\abovedisplayskip{4pt}
  \setlength\belowdisplayskip{3pt}
  \setlength\abovedisplayshortskip{4pt}
  \setlength\belowdisplayshortskip{3pt}
}
\makeatother

\titlespacing{\section}{12pt}{6pt plus 4pt minus 2pt}{6pt plus 4pt minus 2pt}
\titlespacing{\subsection}{6pt}{\parskip}{-\parskip}
\titlespacing{\paragraph}{0pt}{\parskip}{6pt plus 4pt minus 2pt}

\aclfinalcopy % Uncomment this line for the final submission
\def\aclpaperid{910} %  Enter the acl Paper ID here

\newcommand{\V}[1][\mathbf]{#1}
\newcommand{\I}[1][\textit]{#1}
\newcommand{\encstate}[1]{\mathbf{h}_{#1}^{(e)}}
\newcommand{\encstatesec}[1]{\mathbf{h}_{#1}^{(s)}}
\newcommand{\decstate}[1]{\mathbf{h}_{#1}^{(d)}}

\DeclareMathOperator{\linear}{linear}
\DeclareMathOperator{\softmax}{softmax}
\newcommand{\newsoftmax}{\mathop{\mathrm{softmax}}}
\DeclareMathOperator{\score}{score}
\DeclareMathOperator{\RNN}{RNN}
\def\Equal{\texttt{=}}
\title{A Discourse-Aware Attention Model for \\Abstractive Summarization of Long Documents}
  \author{
  {Arman Cohan}$^\dagger$
  \quad {Franck Dernoncourt}$^\star$
  \quad {Doo Soon Kim}$^\star$
  \quad {Trung Bui}$^\star$ \\
  \quad {\textbf{Seokhwan Kim}}$^\star$
  \quad {\textbf{Walter Chang}}$^\star$
  \quad {\textbf{Nazli Goharian}}$^\dagger$ \vspace{6pt} \\
$^\dagger$IRLab, Georgetown University, Washington, DC \\
  {\small \tt \{arman,nazli\}@ir.cs.georgetown.edu} \vspace{4pt}\\
  $^\star$Adobe Research, San Jose, CA\\
  {\small \tt \{dernonco,dkim,bui,seokim,wachang\}@adobe.com}
  }

\date{}
\begin{document}
\maketitle
\begin{abstract}
Neural abstractive summarization models have led to promising results in summarizing relatively short documents. We propose the first model for abstractive summarization of single, longer-form documents (e.g., research papers). Our approach consists of a new hierarchical encoder that models the discourse structure of a document, and an attentive discourse-aware decoder to generate the summary.
Empirical results on two large-scale datasets of scientific papers show that our model significantly outperforms state-of-the-art models.
\end{abstract}

\section{Introduction}

Existing large-scale summarization datasets consist of relatively short documents. For example, articles in the CNN/Daily Mail dataset \cite{hermann2015teaching} are on average about 600 words long. Similarly, existing neural summarization models have focused on summarizing sentences and short documents.
In this work, we propose a model for effective abstractive summarization of longer documents.
Scientific papers are an example of documents that are significantly longer than news articles (see Table \ref{tab:data}). They also follow a standard discourse structure describing the problem, methodology, experiments/results, and finally conclusions \cite{Suppe1998-SUPTSO}.

Most summarization works in the literature focus on extractive summarization. Examples of prominent approaches include frequency-based methods \cite{vanderwende2007beyond}, graph-based methods \cite{erkan2004lexrank}, topic modeling \cite{steinberger2004using}, and neural models \cite{nallapati2017summarunner}. Abstractive summarization is an alternative approach where the generated summary may contain novel words and phrases and is more similar to how humans summarize documents \cite{jing2002using}. Recently, neural methods have led to encouraging results in abstractive summarization \cite{nallapati2016abstractive,see2017get,paulus2017deep,li2017cascaded}. These approaches employ a general framework of sequence-to-sequence (seq2seq) models \cite{sutskever2014sequence} where the document is fed to an encoder network and another (recurrent) network learns to decode the summary. While promising, these methods focus on summarizing news articles which are relatively short. Many other document types, however, are longer and structured. Seq2seq models tend to struggle with longer sequences because at each decoding step, the decoder needs to learn to construct a context vector capturing relevant information from all the tokens in the source sequence \cite{shao2017generating}.

Our main contribution is an abstractive model for summarizing scientific papers which are an example of long-form structured document types. Our model includes a hierarchical encoder, capturing the discourse structure of the document and a discourse-aware decoder that generates the summary. Our decoder attends to different discourse sections and allows the model to more accurately represent important information from the source resulting in a better context vector.
We also introduce two large-scale datasets of long and structured scientific papers obtained from arXiv and PubMed to support both training and evaluating models on the task of long document summarization. Evaluation results show that our method outperforms state-of-the-art summarization models\footnote{\small \urlstyle{rm} Data/code: \url{https://github.com/acohan/long-summarization}}.

\section{Background}

In the seq2seq framework for abstractive summarization, an input document $\V{x}$ is encoded using a Recurrent Neural Network (RNN) with $\encstate{i}$ being the hidden state of the encoder at timestep~$i$. The last step of the encoder is fed as input to another RNN which decodes the output one token at a time. Given an input document along with the corresponding ground-truth summary $\V{y}$, the model is trained to output a summary $\hat{\V{y}}$ that is close to $\V{y}$. The output at timestep $t$ is predicted using the decoder input $\V{x}'_t$, decoder hidden state $\decstate{t}$, and some information about the input sequence. This framework is the general seq2seq framework employed in many generation tasks including machine translation \cite{sutskever2014sequence,bahdanau2014neural} and summarization \cite{nallapati2016abstractive,chopra2016abstractive}.

\begin{figure}[t]
\centering
\includegraphics[width=\linewidth]{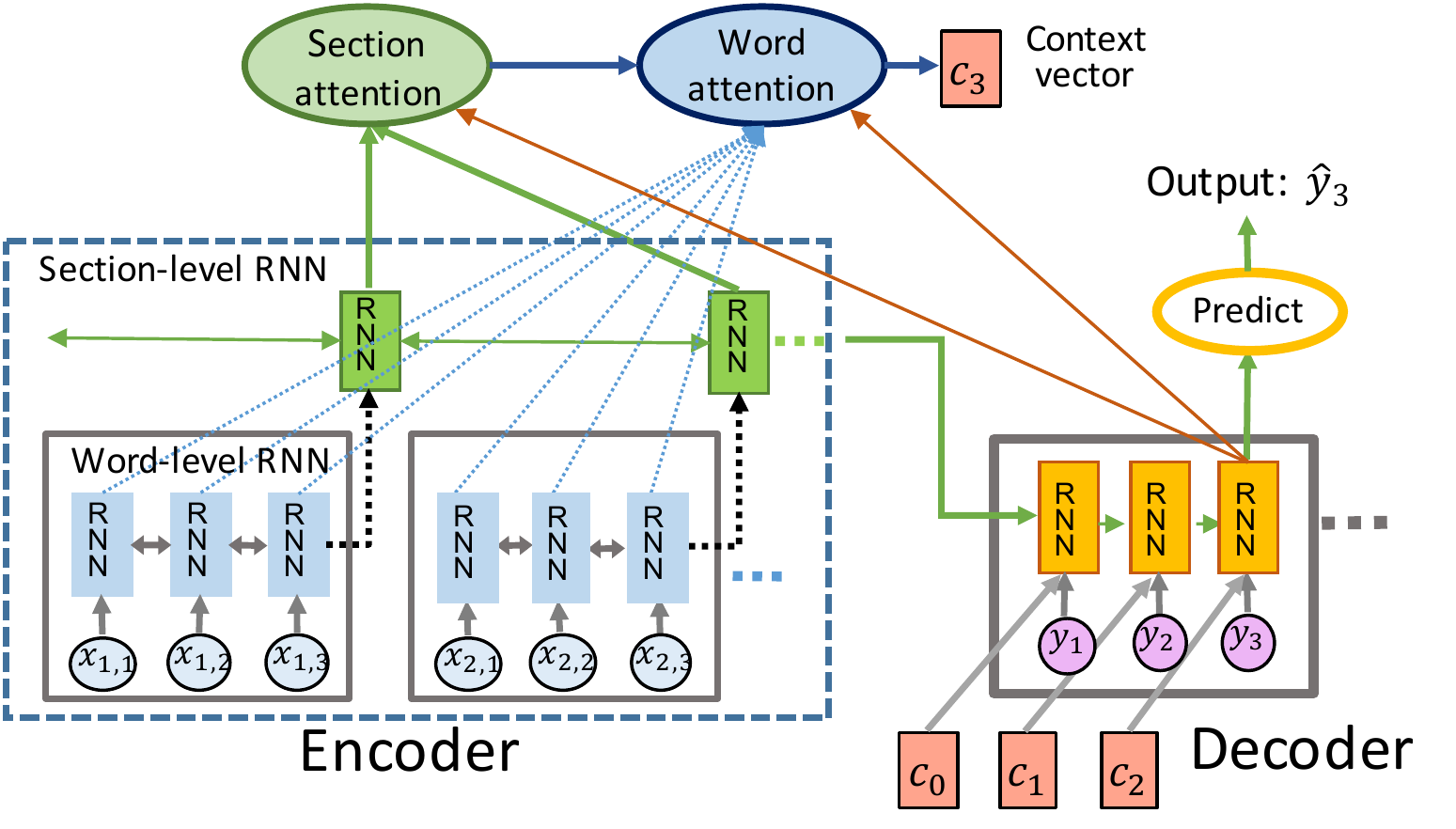}
\caption{Overview of our model.
The word-level RNN is shown in blue and section-level RNN is shown in green. The decoder also consists of an RNN (orange) and a ``predict'' network for generating the summary. At each decoding time step $t$ (here $t$=$3$ is shown), the decoder forms a context vector $c_t$ which encodes the relevant source context ($c_0$ is initialized as a zero vector).
Then the section and word attention weights are respectively computed using the green ``section attention'' and the blue ``word attention'' blocks.
The context vector is used as another input to the decoder RNN and as an input to the ``predict'' network which outputs the next word using a joint pointer-generator network.
}
\label{fig:arch}
\end{figure}

\paragraph{Attentive decoding}

The attention mechanism maps the decoder state and the encoder states to an output vector, which is a weighted sum of the encoder states and is called context vector \cite{bahdanau2014neural}. Incorporating this context vector at each decoding timestep (attentive decoding) is proven effective in seq2seq models.
Formally, the context vector $c_t$ is defined as: $ \V{c}_t \!=\! \sum_{i=1}^{N}\alpha^{(t)}_i \encstate{i} $
where $\alpha^{(t)}_i$ are the attention weights calculated as follows:
\begin{equation}\small \alpha^{(t)}_i \!=\! \newsoftmax_{i}(\score(\encstate{i}, \decstate{t-1})) \end{equation}

\noindent where {\small$\newsoftmax\limits_{i}$} means that the denominator's sum in the softmax function is over $i$. The $\score$ function can be defined in bilinear, additive, or multiplicative ways \cite{luong2015effective}. We use the additive scoring function:
\begin{equation}
  \small
  \score(\encstate{i}, \decstate{t-1}) = \V{v}_a^\top \tanh \big( \linear(\encstate{i}, \decstate{t-1}) \big) \label{eq:score-attn}
\end{equation}
\noindent where $\V{v}_a$ is a weight vector and $\linear$ is a linear mapping function. I.e.,
\begin{equation}
\small
\linear(\contour[2]{black}{\textsc{x}}_1, \contour[2]{black}{\textsc{x}}_2) = \contour[2]{black}{\textsc{w}}_1 \contour[2]{black}{\textsc{x}}_1 + \contour[2]{black}{\textsc{w}}_2 \contour[2]{black}{\textsc{x}}_2 + \V{b}
\end{equation}
\noindent where \contour[2]{black}{\textsc{w}}$_1$ and \contour[2]{black}{\textsc{w}}$_2$ are weight matrices and $\V{b}$ is the bias vector.

\section{Model}

We now describe our discourse-aware summarization model (shown in Figure \ref{fig:arch}).

\paragraph{Encoder}

Our encoder extends the RNN encoder to a hierarchical RNN that captures the document discourse structure. We first encode each discourse section and then encode the document. Formally, we encode the document as a vector $\V{d}$ according to the following:
$$\V{d}=\RNN_{doc} \big( \{\encstatesec{1}, ..., \encstatesec{N}\} \big)$$

\noindent $\RNN(.)$ denotes a function which is a recurrent neural network whose output is the final state of the network encoding the entire sequence. $N$ is the number of sections in the document and $\encstatesec{j}$ is representation of section $j$ in the document consisting of a sequence of tokens.
$$\encstatesec{j} = \RNN_{sec} \big( \V{x}_{(j,1)}, ... \V{x}_{(j,M)}\}\big)$$

\noindent where $\V{x}_{(j,i)}$ are dense embeddings corresponding to the tokens $w_{(j,i)}$ and $M$ is the maximum section length. The parameters of $\RNN_{sec}$ are shared for all the discourse sections. We use a single layer bidirectional LSTM (following the LSTM formulation of \newcite{graves2013speech}) for both $\RNN_{doc}$ and $\RNN_{sec}$; further extension to multilayer LSTMs is straightforward. We combine the forward and backward LSTM states to a single state using a simple feed-forward network:
$$\V{h}=\mathrm{relu}(\V{W}([\overrightarrow{\V{h}},\overleftarrow{\V{h}}]+\V{b})$$
where $[,]$ shows the concatenation operation.
Throughout, when we mention the RNN (LSTM) state, we are referring to this combined state of both forward and backward RNNs (LSTMs).

\paragraph{Discourse-aware decoder}

When humans summarize a long structured document, depending on the domain and the nature of the document, they write about important points from different discourse sections of the document. For example, scientific paper abstracts typically include the description of the problem, discussion of the methods, and finally results and conclusions \cite{Suppe1998-SUPTSO}. Motivated by this observation, we propose a discourse-aware attention method.
Intuitively, at each decoding timestep, in addition to the words in the document, we also attend to the relevant discourse section (the ``section attention'' block in Figure \ref{fig:arch}). Then we use the discourse-related information to modify the word-level attention function. Specifically, the context vector representing the source document is:
\begin{equation}
\V{c}_t = \sum\nolimits_{j=1}^N \sum\nolimits_{i=1}^M \alpha_{(j,i)}^{(t)} \encstate{(j,i)}
\label{eq:context} \end{equation}

\noindent where $\encstate{(j,i)}$ shows the encoder state of word $i$ in discourse section $j$ and $\alpha_{(j,i)}^{(t)}$ shows the corresponding attention weight to that encoder state. The scalar weights $\alpha_{(j,i)}^{(t)}$ are obtained according to:

\begin{equation} \alpha^{(t)}_{(j,i)} = \newsoftmax_{(i,j)} \Big( \beta^{(t)}_j \score(\encstate{(j,i)}, \decstate{t-1}) \Big) \label{eq:att-word}\end{equation}

\noindent The $\score$ function is the additive attention function (Equation \ref{eq:score-attn}) and the weights $\beta^{(t)}_j$ are updated according to:
\begin{equation}
\beta^{(t)}_{j} = \newsoftmax_j(\score(\encstatesec{j}, \decstate{t-1})) \label{eq:att-sec}
\end{equation}

At each timestep $t$, the decoder state $\decstate{t}$ and the context vector $\V{c}_t$ are used to estimate the probability distribution of next word $y_t$:
\begin{equation}
  \small
  \label{eq:final-softmax}
  p(y_t|y_{1:t-1}) = \softmax \big(\V{V}^\top \linear \big( \decstate{t}, \V{c}_t \big)\big)
\end{equation}
% \vspace{-10pt}

\noindent where $\V{V}$ is a vocabulary weight matrix and $\softmax$ is over the entire vocabulary.

\paragraph{Copying from source}

There has been a surge of recent works in sequence learning tasks to address the problem of \I{unkown} token prediction by allowing the model to occasionally copy words directly from source instead of generating a new token \cite{gu2016incorporating,see2017get,paulus2017deep,wiseman2017challenges}. Following these works, we add an additional binary variable $z_t$ to the decoder, indicating generating a word from vocabulary ($z_t\Equal0$) or copying a word from the source ($z_t\Equal1$). The probability is learnt during training according to the following equation:
\begin{equation} p(z_t\Equal1|y_{1:t-1})=\sigma(\linear(\decstate{t}, \V{c}_t, \V{x}'_t ))
\label{eq:switch} \end{equation}

Then the next word $y_t$ is generated according to:
$$ p(y_t|y_{1:t-1}) = \sum_z p(y_t, z_t\Equal z | y_{1:t-1}) ; z=\{0,1\} $$

The joint probability is decomposed as:
\vspace{-10pt}

{\small
\begin{align*}
  & p(y_t, z_t\Equal z) =
\begin{cases}
  p_{c}(y_t| y_{1:t-1}) \; p(z_t\Equal z|y_{1:t-1}), \quad z\Equal1 \\
  p_{g}(y_t| y_{1:t-1}) \; p(z_t\Equal  z|y_{1:t-1}), \quad z\Equal0
\end{cases}
\end{align*}
}

\noindent $p_g$ is the probability of generating a word from the vocabulary and is defined according to Equation~\ref{eq:final-softmax}.
$p_c$ is the probability of copying a word from the source vector $\V{x}$ and is defined as the sum of the word's attention weights. Specifically, the probability of copying a word $x_\ell$ is defined as:
{
\begin{equation}  p_{c}(y_t=x_\ell| y_{1:t-1}) = \sum\nolimits_{(j,i):x_{(j,i)}=x_\ell}\alpha^{(t)}_{(j,i)} \vspace{-9pt} \label{eq:pointer}
\end{equation}}

\paragraph{Decoder coverage}
In long sequences, the neural generation models tend to repeat phrases where the softmax layer predicts the same phrase multiple times over multiple timesteps.
To address this issue, following \newcite{see2017get}, we track attention coverage to avoid repeatedly attending to the same steps. This is done with a coverage vector $\V{cov}^{(t)}$, the sum of attention weight vectors at previous timesteps:
% {\small $\V{cov}^{(t)} = \sum_{k=0}^{t-1} \boldsymbol{\alpha}^{(k)}$}
{\small $\mathrm{cov}^{(t)}_{(j,i)} = \sum_{k=0}^{t-1} \alpha^{(k)}_{(j,i)}$}

% \noindent where $\boldsymbol{\alpha}^{(k)}$ is the vector of attention weights at timestep $k$.
The coverage implicitly includes information about the attended document discourse sections. We incorporate the decoder coverage as an additional input to the attention function:

{\small
\begin{equation*}
\alpha^{(t)}_{(j,i)} = \newsoftmax_{(i,j)} \Big( \beta^{(t)}_j \score(\encstate{(j,i)}, \mathrm{cov}^{(t)}_{(j,i)} ,\decstate{t-1}) \Big) \label{eq:cov}
\end{equation*}
\vspace{-10pt}}
\normalsize

\section{Related work}
Neural abstractive summarization models have been studied in the past \cite{rush2015neural,chopra2016abstractive,nallapati2016abstractive} and later extended by source copying  \cite{miao2016language,see2017get}, reinformcement learning \cite{paulus2017deep}, and sentence salience information \cite{li2017cascaded}. One model variant of \newcite{nallapati2016abstractive} is related to our model in using sentence-level information in attention. However, our model is different as it contains a hierarchical encoder, uses discourse
sections in the decoding step, and has a coverage mechanism. Similarly, \citet{ling-rush:2017:FrontiersSummarization} proposed a coarse-to-fine attention model that uses hard attention to find the text chunks of importance and then only attend to words in that chunk. In contrast, we consider all the discourse sections using soft attention. The closest model to ours is that of \newcite{see2017get} and \newcite{paulus2017deep} who used a joint pointer-generator network for summarization. However, our model extends theirs by \textit{(i)} a hierarchical encoder for modeling long documents and \textit{(ii)} a discourse-aware decoder that captures the information flow from all discourse sections of the document. Finally, in a recent work, \citet{j.2018generating} proposed a model based on the transformer network \cite{vaswani2017attention} for abstractive generation of Wikipedia articles. However, their focus is on multi-document summarization.

Our datasets are obtained from scientific papers. Scientific document summarization has been recently received extended attention \cite{qazvinian2013generating,cohan-goharian:2015:EMNLP,Cohan2017,cohan2017contextualizing}. In contrast to ours, existing approaches are extractive and rely on external information such as citations, which may not be available for all papers.

\section{Data}
Seq2seq models typically have a large number of parameters and thus they require large training data with ground truth summaries. Researchers have constructed such training data from news articles (e.g., CNN, Daily Mail and New York Times articles), where the abstracts or highlights of news articles are considered as ground truth summaries \cite{nallapati2016abstractive,paulus2017deep}. However, news articles are relatively short and not suitable for the task of long-from document summarization.
Following these works, we take scientific papers as an example of long documents with discourse information, where their abstracts can be used as ground-truth summaries. We introduce two datasets collected from scientific repositories, arXiv.org and PubMed.com.

\begin{table}[]
\centering
\small
\setlength{\tabcolsep}{3pt}
\begin{tabular}{@{}lrrr@{}}
\toprule
Datasets           & \# docs & \begin{tabular}[c]{@{}r@{}}avg. doc.\\ length (words)\end{tabular} & \begin{tabular}[c]{@{}r@{}}avg. summary\\ length (words)\end{tabular} \\ \midrule
CNN                & 92K     & 656                                                       & 43                                                          \\
Daily Mail         & 219K    & 693                                                       & 52                                                          \\
NY Times          & 655K    & 530                                                       & 38                                                          \\
PubMed (this work) & 133K    & 3016                                                          &  203                                                           \\
arXiv (this work)  & 215K    & 4938                                                      & 220                                                         \\ \bottomrule
\end{tabular}
\caption{Statistics of our arXiv and PubMed datasets compared with existing large-scale summarization corpora, CNN and Daily Mail \cite{nallapati2016abstractive} and NY Times \cite{paulus2017deep}.}
\label{tab:data}
\vspace{-6pt}
\end{table}

The choice of scientific papers for our dataset is motivated by the fact that scientific papers are examples of long documents that follow a standard discourse structure and they already come with ground truth summaries, making it possible to train supervised neural models. We follow existing work in constructing large-scale summarization datasets that take news article abstracts as ground truth.

We remove the documents that are excessively long (e.g., theses) or too short (e.g., tutorial announcements), or do not have an abstract or discourse structure. We use the level-1 section headings as the discourse information. For arXiv, we use the \LaTeX\xspace files and convert them to plain text using Pandoc (https://pandoc.org) to preserve the discourse section information. We remove figures and tables using regular expressions to only preserve the textual information. We also normalize math formulas and citation markers with special tokens. We analyze the document section names and identify the most common concluding sections names (e.g. \I{conclusion}, \I{concluding remarks}, \I{summary}, etc). We only keep the sections up to the conclusion section of the document and we remove sections after the conclusion.

The statistics of our datasets are shown in Table \ref{tab:data}. In our datasets, both document and summary lengths are significantly larger than the existing large-scale summarization datasets. We retain about 3\%  (5\%) of PubMed (ArXiv) as validation data and about another 3\% (5\%) for test; the rest is used for training.

\section{Experiments}
\paragraph{Setup}

Similar to the majority of published research in the summarization literature \cite{chopra2016abstractive,nallapati2016abstractive,see2017get}, evaluation was done using the \textsc{Rouge} automatic summarization evaluation metric \cite{lin2004rouge} with full-length F-1 \textsc{Rouge} scores. We lowercase all tokens and perform sentence and word tokenization using spaCy \cite{honnibal-johnson:2015:EMNLP}.
% We evaluate our models on the arXiv dataset described in the previous section.

\paragraph{Implementation details}
We use Tensorflow 1.4 for implementing our models. We use the hyperparameters suggested by \citet{see2017get}. In particular, we use two bidirectional LSTMs with cell size of 256 and embedding dimensions of 128. Embeddings are trained from scratch and we did not find any gain using pre-trained embeddings. The vocabulary size is constrained to 50,000; using larger vocabulary size did not result in any improvement. We use mini-batches of size 16 and we limit the document length to 2000 and section length to 500 tokens, and number of sections to 4. We use batch-padding and dynamic unrolling to handle variable sequence lengths in LSTMs. Training was done using Adagrad optimizer with learning rate 0.15
and an initial accumulator value of 0.1. The maximum decoder size was 210 tokens which is in line with average abstract length in our datasets. We first train the model without coverage and added it at the last two epochs to help the model converge faster. We train the models on NVIDIA Titan X Pascal GPUs. Training is performed for about 10 epochs and each training step takes about 3.2 seconds. We used beam search at decoding time with beam size of 4. We train the abstractive baselines for about 250K iterations as suggested by their authors.

\paragraph{Comparison}
We compare our method with several well-known extractive baselines as well as state-of-the-art abstractive models using their open-sourced implementations, when available; we follow the same training setup described in the corresponding papers. The compared methods are:
  \textit{LexRank} \cite{erkan2004lexrank},
  \textit{SumBasic} \cite{vanderwende2007beyond},
  \textit{LSA} \cite{steinberger2004using},
  \textit{Attn-Seq2Seq} \cite{nallapati2016abstractive,chopra2016abstractive},
  \textit{Pntr-Gen-Seq2Seq} \cite{see2017get}. The first three are extractive models and last two are abstractive. \textit{Pntr-Gen-Seq2Seq} extends \textit{Attn-Seq2Seq} by using a joint pointer network during decoding.
  For \textit{Pntr-Gen-Seq2Seq} we use their reported hyperparameters to ensure that the result differences are not due to hyperparameter tuning.

% \paragraph{pre-processing}

\begin{table}[]
\centering
\small
\setlength{\tabcolsep}{4pt}
\begin{tabular}{@{}llrrrr@{}}
\toprule
 \multicolumn{2}{@{}l}{Summarizer}         & RG-1            & RG-2            & RG-3            & RG-L            \\ \midrule
\multirow{3}{*}{\rotatebox[origin=c]{90}{\tiny{Extractive}}}
& SumBasic           & 29.47  & 6.95  & 2.36  & 26.30  \\
& LexRank          & 33.85  & 10.73  & \bf{4.54}  & 28.99  \\
& LSA         & 29.91  & 7.42  & 3.12  & 25.67  \\ \specialrule{0.1pt}{0.5pt}{0.5pt}
\multirow{3}{*}{\rotatebox[origin=c]{90}{\tiny{Abstractive}}}
& Attn-Seq2Seq     & 29.30  & 6.00  & 1.77  & 25.56  \\
& Pntr-Gen-Seq2Seq & 32.06     & 9.04         & 2.15        & 25.16     \\
& This work  &  $^\dagger$$^\ddagger$\bf{35.80}  &  $^\dagger$\bf{11.05}  &  $^\dagger$3.62  &  $^\dagger$$^\ddagger$\bf{31.80}  \\
\bottomrule
\end{tabular}
\caption{\small{Results on the arXiv dataset,  RG: \textsc{Rouge}. For our method $^\dagger$~($^\ddagger$) shows statistically significant improvement with $p{<}0.05$ over
 other abstractive methods (all other methods).}}
\label{tab:resarXiv}
\end{table}

\begin{table}[]
\centering
\setlength{\tabcolsep}{4pt}
\small
\begin{tabular}{@{}llrrrr@{}}
\toprule
 \multicolumn{2}{@{}l}{Summarizer}      & RG-1  & RG-2  & RG-3 & RG-L  \\ \midrule
\multirow{3}{*}{\rotatebox[origin=c]{90}{\tiny{Extractive}}}
& SumBasic         & 37.15 & 11.36 & 5.42 & 33.43 \\
& LexRank          & \bf{39.19} & 13.89 & 7.27 & 34.59 \\
& LSA              & 33.89 & 9.93 & 5.04 & 29.70 \\ \specialrule{0.1pt}{0.5pt}{0.5pt}
\multirow{3}{*}{\rotatebox[origin=c]{90}{\tiny{Abstractive}}}
& Attn-Seq2Seq     & 31.55 & 8.52  & 7.05 & 27.38 \\
& Pntr-Gen-Seq2Seq & 35.86    & 10.22    & 7.60   & 29.69    \\
& This work        & $^\dagger$38.93    & $^\dagger$$^\ddagger$\bf{15.37}    & $^\dagger$$^\ddagger$\bf{9.97}   & $^\dagger$$^\ddagger$\bf{35.21}    \\ \bottomrule
\end{tabular}
\caption{\small{Results on PubMed dataset, RG:\textsc{Rouge}. For our method, $^\dagger$~($^\ddagger$) shows statistically significant improvement with $p{<}0.05$ over abstractive methods (all other methods).}}
\label{tab:resPubMed}
\vspace{-6pt}
\end{table}

\paragraph{Results}
Our main results are shown in Tables~\ref{tab:resarXiv} and \ref{tab:resPubMed}. Our model significantly outperforms the state-of-the-art abstractive methods, showing its effectiveness on both datasets. We observe that in our \textsc{Rouge}-1 score is respectively about 4 and 3 points higher than the abstractive model \textit{Pntr-Gen-Seq2Seq} for the arXiv and PubMed datasets, providing a significant improvement. Our method also outperforms most of the extractive methods except for \textit{LexRank} in one of the \textsc{Rouge} scores. We note that since extractive methods copy salient sentences from the document, it is usually easier for them to achieve higher \textsc{Rouge} scores.

Figure \ref{fig:example} illustrates the effectiveness of our model extensions in capturing various discourse information from the papers. It can be observed that the state-of-the-art \textit{Pntr-Gen-Seq2Seq} model generates a summary that mostly focuses on introducing the problem, whereas our model generates a summary that includes more information about the methodology and impacts of the target paper. This indicates that the context vector in our model compared with \textit{Pntr-Gen-Seq2Seq} is better able to capture important information from the source by attending to various discourse sections.

\begin{figure}[]
\centering
\scriptsize
\renewcommand*{\arraystretch}{1.3}
\begin{tabular}{@{}p{\linewidth}@{}}
\toprule
\textit{Abstract}: in this paper , the author proposes a series of multilevel double hashing schemes called cascade hash tables . they use several levels of hash tables . in each table , we use the common double hashing scheme . higher level hash tables work as fail - safes of lower level hash tables . by this strategy , it could effectively reduce collisions in hash insertion . thus it gains a constant worst case lookup time with a relatively high load factor (@xmath0 ) in random experiments . different parameters of cascade hash tables are tested .
\\
\textit{Pntr-Gen-Seq2Seq}: hash table is a common data structure used in large set of data storage and retrieval . it has an o(1 ) lookup time on average , but the worst case lookup time can be as bad as . is the size of the hash table . we present a set of hash table schemes called cascade hash tables . hash table data structures which consist of several of hash tables with different size .
\\
\textit{Our method}: cascade hash tables are a common data structure used in large set of data storage and retrieval . such a time variation is essentially caused by possibly many collisions during keys hashing . in this paper , we present a set of hash schemes called cascade hash tables which consist of several levels ( @xmath2 ) of hash tables with different size . after constant probes , if an item ca 'nt find a free slot in limited probes in any hash table , it will try to find a cell in the second level , or subsequent lower levels . with this simple strategy , these hash tables will have descendant load factors , therefore lower collision probabilities . \\ \bottomrule
\end{tabular}
\caption{Example of a generated summary}
\label{fig:example}
\vspace{-2pt}
\end{figure}

\section{Conclusions and future work}
This work was the first attempt at addressing neural abstractive summarization of single, long documents. We presented a neural sequence-to-sequence model that is able to effectively summarize long and structured documents such as scientific papers. While our results are encouraging, there is still much room for improvement for this challenging task; our new datasets can help the community to further explore this problem.

We note that following the convention in the summarization research, our quantitative evaluation is performed by \textsc{Rouge} automatic metric. While \textsc{Rouge} is an effective evaluation framework, nuances in the coherence or coverage of the summaries are not captured with it. It is non-trivial to evaluate such qualities especially for long document summarization; future work can design expert human evaluations to explore these nuances.

\section*{Acknowledgements}
We thank the three anonymous reviewers for their comments and suggestions.

\bigskip

\bibliography{naaclhlt2018}

\begin{thebibliography}{}
\expandafter\ifx\csname natexlab\endcsname\relax\def\natexlab#1{#1}\fi

\bibitem[{Bahdanau et~al.(2014)Bahdanau, Cho, and Bengio}]{bahdanau2014neural}
Dzmitry Bahdanau, Kyunghyun Cho, and Yoshua Bengio. 2014.
\newblock Neural machine translation by jointly learning to align and
  translate.
\newblock {\em arXiv preprint arXiv:1409.0473\/} .

\bibitem[{Chopra et~al.(2016)Chopra, Auli, Rush, and
  Harvard}]{chopra2016abstractive}
Sumit Chopra, Michael Auli, Alexander~M Rush, and SEAS Harvard. 2016.
\newblock Abstractive sentence summarization with attentive recurrent neural
  networks.
\newblock In {\em HLT-NAACL\/}. pages 93--98.

\bibitem[{Cohan and Goharian(2015)}]{cohan-goharian:2015:EMNLP}
Arman Cohan and Nazli Goharian. 2015.
\newblock \href{http://aclweb.org/anthology/D15-1045}{Scientific article
  summarization using citation-context and article's discourse structure}.
\newblock In {\em Proceedings of the 2015 Conference on Empirical Methods in
  Natural Language Processing\/}. Association for Computational Linguistics,
  Lisbon, Portugal, pages 390--400.
\newblock \url{http://aclweb.org/anthology/D15-1045}.

\bibitem[{Cohan and Goharian(2017{\natexlab{a}})}]{cohan2017contextualizing}
Arman Cohan and Nazli Goharian. 2017{\natexlab{a}}.
\newblock Contextualizing citations for scientific summarization using word
  embeddings and domain knowledge.
\newblock {\em arXiv preprint arXiv:1705.08063\/} .

\bibitem[{Cohan and Goharian(2017{\natexlab{b}})}]{Cohan2017}
Arman Cohan and Nazli Goharian. 2017{\natexlab{b}}.
\newblock \href{https://doi.org/10.1007/s00799-017-0216-8}{Scientific document
  summarization via citation contextualization and scientific discourse}.
\newblock {\em International Journal on Digital Libraries\/} pages 1--17.
\newblock \url{https://doi.org/10.1007/s00799-017-0216-8}.

\bibitem[{Erkan and Radev(2004)}]{erkan2004lexrank}
G{\"u}nes Erkan and Dragomir~R Radev. 2004.
\newblock Lexrank: Graph-based lexical centrality as salience in text
  summarization.
\newblock {\em Journal of Artificial Intelligence Research\/} 22:457--479.

\bibitem[{Graves et~al.(2013)Graves, Mohamed, and Hinton}]{graves2013speech}
Alex Graves, Abdel-rahman Mohamed, and Geoffrey Hinton. 2013.
\newblock Speech recognition with deep recurrent neural networks.
\newblock In {\em Acoustics, speech and signal processing (icassp), 2013 ieee
  international conference on\/}. IEEE, pages 6645--6649.

\bibitem[{Gu et~al.(2016)Gu, Lu, Li, and Li}]{gu2016incorporating}
Jiatao Gu, Zhengdong Lu, Hang Li, and Victor~O.K. Li. 2016.
\newblock \href{http://www.aclweb.org/anthology/P16-1154}{Incorporating copying
  mechanism in sequence-to-sequence learning}.
\newblock In {\em Proceedings of the 54th Annual Meeting of the Association for
  Computational Linguistics (Volume 1: Long Papers)\/}. Association for
  Computational Linguistics, Berlin, Germany, pages 1631--1640.
\newblock \url{http://www.aclweb.org/anthology/P16-1154}.

\bibitem[{Hermann et~al.(2015)Hermann, Kocisky, Grefenstette, Espeholt, Kay,
  Suleyman, and Blunsom}]{hermann2015teaching}
Karl~Moritz Hermann, Tomas Kocisky, Edward Grefenstette, Lasse Espeholt, Will
  Kay, Mustafa Suleyman, and Phil Blunsom. 2015.
\newblock Teaching machines to read and comprehend.
\newblock In {\em Advances in Neural Information Processing Systems\/}. pages
  1693--1701.

\bibitem[{Honnibal and Johnson(2015)}]{honnibal-johnson:2015:EMNLP}
Matthew Honnibal and Mark Johnson. 2015.
\newblock \href{https://aclweb.org/anthology/D/D15/D15-1162}{An improved
  non-monotonic transition system for dependency parsing}.
\newblock In {\em Proceedings of the 2015 Conference on Empirical Methods in
  Natural Language Processing\/}. Association for Computational Linguistics,
  Lisbon, Portugal, pages 1373--1378.
\newblock \url{https://aclweb.org/anthology/D/D15/D15-1162}.

\bibitem[{Jing(2002)}]{jing2002using}
Hongyan Jing. 2002.
\newblock Using hidden markov modeling to decompose human-written summaries.
\newblock {\em Computational linguistics\/} 28(4):527--543.

\bibitem[{Li et~al.(2017)Li, Lam, Bing, Guo, and Li}]{li2017cascaded}
Piji Li, Wai Lam, Lidong Bing, Weiwei Guo, and Hang Li. 2017.
\newblock Cascaded attention based unsupervised information distillation for
  compressive summarization.
\newblock In {\em Proceedings of the 2017 Conference on Empirical Methods in
  Natural Language Processing\/}. pages 2071--2080.

\bibitem[{Lin(2004)}]{lin2004rouge}
Chin-Yew Lin. 2004.
\newblock Rouge: A package for automatic evaluation of summaries.
\newblock In {\em Text summarization branches out: Proceedings of the ACL-04
  workshop\/}. Barcelona, Spain, volume~8.

\bibitem[{Ling and Rush(2017)}]{ling-rush:2017:FrontiersSummarization}
Jeffrey Ling and Alexander Rush. 2017.
\newblock \href{http://www.aclweb.org/anthology/W17-4505}{Coarse-to-fine
  attention models for document summarization}.
\newblock In {\em Proceedings of the Workshop on New Frontiers in
  Summarization\/}. Association for Computational Linguistics, Copenhagen,
  Denmark, pages 33--42.
\newblock \url{http://www.aclweb.org/anthology/W17-4505}.

\bibitem[{Liu et~al.(2018)Liu, Saleh, Pot, Goodrich, Sepassi, Kaiser, and
  Shazeer}]{j.2018generating}
Peter~J. Liu, Mohammad Saleh, Etienne Pot, Ben Goodrich, Ryan Sepassi, Lukasz
  Kaiser, and Noam Shazeer. 2018.
\newblock \href{https://openreview.net/forum?id=Hyg0vbWC-}{Generating wikipedia
  by summarizing long sequences}.
\newblock In {\em International Conference on Learning Representations\/}.
\newblock \url{https://openreview.net/forum?id=Hyg0vbWC-}.

\bibitem[{Luong et~al.(2015)Luong, Pham, and Manning}]{luong2015effective}
Minh-Thang Luong, Hieu Pham, and Christopher~D Manning. 2015.
\newblock Effective approaches to attention-based neural machine translation.
\newblock {\em arXiv preprint arXiv:1508.04025\/} .

\bibitem[{Miao and Blunsom(2016)}]{miao2016language}
Yishu Miao and Phil Blunsom. 2016.
\newblock Language as a latent variable: Discrete generative models for
  sentence compression.
\newblock {\em arXiv preprint arXiv:1609.07317\/} .

\bibitem[{Nallapati et~al.(2017)Nallapati, Zhai, and
  Zhou}]{nallapati2017summarunner}
Ramesh Nallapati, Feifei Zhai, and Bowen Zhou. 2017.
\newblock Summarunner: A recurrent neural network based sequence model for
  extractive summarization of documents.
\newblock {\em AAAI\/} 1:1.

\bibitem[{Nallapati et~al.(2016)Nallapati, Zhou, Gulcehre, Xiang
  et~al.}]{nallapati2016abstractive}
Ramesh Nallapati, Bowen Zhou, Caglar Gulcehre, Bing Xiang, et~al. 2016.
\newblock Abstractive text summarization using sequence-to-sequence rnns and
  beyond.
\newblock {\em arXiv preprint arXiv:1602.06023\/} .

\bibitem[{Paulus et~al.(2017)Paulus, Xiong, and Socher}]{paulus2017deep}
Romain Paulus, Caiming Xiong, and Richard Socher. 2017.
\newblock A deep reinforced model for abstractive summarization.
\newblock {\em arXiv preprint arXiv:1705.04304\/} .

\bibitem[{Qazvinian et~al.(2013)Qazvinian, Radev, Mohammad, Dorr, Zajic,
  Whidby, and Moon}]{qazvinian2013generating}
Vahed Qazvinian, Dragomir~R Radev, Saif~M Mohammad, Bonnie Dorr, David Zajic,
  Michael Whidby, and Taesun Moon. 2013.
\newblock Generating extractive summaries of scientific paradigms.
\newblock {\em Journal of Artificial Intelligence Research\/} 46:165--201.

\bibitem[{Rush et~al.(2015)Rush, Chopra, and Weston}]{rush2015neural}
Alexander~M Rush, Sumit Chopra, and Jason Weston. 2015.
\newblock A neural attention model for abstractive sentence summarization.
\newblock {\em arXiv preprint arXiv:1509.00685\/} .

\bibitem[{See et~al.(2017)See, Manning, and Liu}]{see2017get}
Abigail See, Christopher Manning, and Peter Liu. 2017.
\newblock \href{https://arxiv.org/abs/1704.04368}{Get to the point:
  Summarization with pointer-generator networks}.
\newblock In {\em Association for Computational Linguistics\/}.
\newblock \url{https://arxiv.org/abs/1704.04368}.

\bibitem[{Shao et~al.(2017)Shao, Gouws, Britz, Goldie, Strope, and
  Kurzweil}]{shao2017generating}
Yuanlong Shao, Stephan Gouws, Denny Britz, Anna Goldie, Brian Strope, and Ray
  Kurzweil. 2017.
\newblock Generating high-quality and informative conversation responses with
  sequence-to-sequence models.
\newblock In {\em Proceedings of the 2017 Conference on Empirical Methods in
  Natural Language Processing\/}. pages 2210--2219.

\bibitem[{Steinberger and Jezek(2004)}]{steinberger2004using}
Josef Steinberger and Karel Jezek. 2004.
\newblock Using latent semantic analysis in text summarization and summary
  evaluation.
\newblock In {\em Proc. ISIM’04\/}. pages 93--100.

\bibitem[{Suppe(1998)}]{Suppe1998-SUPTSO}
Frederick Suppe. 1998.
\newblock The structure of a scientific paper.
\newblock {\em Philosophy of Science\/} 65(3):381--405.

\bibitem[{Sutskever et~al.(2014)Sutskever, Vinyals, and
  Le}]{sutskever2014sequence}
Ilya Sutskever, Oriol Vinyals, and Quoc~V Le. 2014.
\newblock Sequence to sequence learning with neural networks.
\newblock In {\em Advances in neural information processing systems\/}. pages
  3104--3112.

\bibitem[{Vanderwende et~al.(2007)Vanderwende, Suzuki, Brockett, and
  Nenkova}]{vanderwende2007beyond}
Lucy Vanderwende, Hisami Suzuki, Chris Brockett, and Ani Nenkova. 2007.
\newblock Beyond sumbasic: Task-focused summarization with sentence
  simplification and lexical expansion.
\newblock {\em Information Processing \& Management\/} 43(6):1606--1618.

\bibitem[{Vaswani et~al.(2017)Vaswani, Shazeer, Parmar, Uszkoreit, Jones,
  Gomez, Kaiser, and Polosukhin}]{vaswani2017attention}
Ashish Vaswani, Noam Shazeer, Niki Parmar, Jakob Uszkoreit, Llion Jones,
  Aidan~N Gomez, \L~ukasz Kaiser, and Illia Polosukhin. 2017.
\newblock
  \href{http://papers.nips.cc/paper/7181-attention-is-all-you-need.pdf}{Attention
  is all you need}.
\newblock In I.~Guyon, U.~V. Luxburg, S.~Bengio, H.~Wallach, R.~Fergus,
  S.~Vishwanathan, and R.~Garnett, editors, {\em Advances in Neural Information
  Processing Systems 30\/}, Curran Associates, Inc., pages 6000--6010.
\newblock \url{http://papers.nips.cc/paper/7181-attention-is-all-you-need.pdf}.

\bibitem[{Wiseman et~al.(2017)Wiseman, Shieber, and
  Rush}]{wiseman2017challenges}
Sam Wiseman, Stuart~M Shieber, and Alexander~M Rush. 2017.
\newblock Challenges in data-to-document generation.
\newblock {\em arXiv preprint arXiv:1707.08052\/} .

\end{thebibliography}
\bibliographystyle{acl_natbib}

% \appendix

% \section{Dataset construction}
% \label{sec:data1}

% \section{Multiple Appendices}
% \dots can be gotten by using more than one section. We hope you won't
% need that.

\end{document}